\pdfoutput=1

\documentclass[11pt]{article}

\usepackage[]{acl}

\usepackage{times}
\usepackage{latexsym}

\usepackage[T1]{fontenc}

\usepackage[utf8]{inputenc}
\usepackage{graphicx}
\usepackage{amsmath}
\usepackage{amsfonts}
\usepackage{microtype}
\usepackage{floatrow}
\usepackage{caption}
\usepackage{multirow}
\usepackage{booktabs}
\usepackage{subcaption}

\newcommand{\model}[1]{{TReasoner}} 
\newcommand{\mydataset}[1]{{CondNLI}} 

\title{Reasoning over Logically Interacted Conditions for Question Answering}

\author{Haitian Sun \\
  School of Computer Science \\
  Carnegie Mellon University \\
  \texttt{haitians@cs.cmu.edu} \\\And
  William W. Cohen \\
  Google Research \\
  \texttt{wcohen@google.com} \\\And 
  Ruslan Salakhutdinov \\
  School of Computer Science \\
  Carnegie Mellon University \\
  \texttt{rsalakhu@cs.cmu.edu} \\}

\begin{document}
\maketitle

\begin{abstract}

Some questions have multiple answers that are not equally correct, i.e. answers are different under different conditions. Conditions are used to distinguish answers as well as to provide additional information to support them. 
In this paper, we study a more challenging task where answers are constrained by a list of conditions that logically interact, which requires performing logical reasoning over the conditions to determine the correctness of the answers. Even more challenging, we only provide evidences for a subset of the conditions, so some questions may not have deterministic answers. In such cases, models are asked to find probable answers and identify conditions that need to be satisfied to make the answers correct. We propose a new model, \model{}, for this challenging reasoning task. \model{} consists of an entailment module, a reasoning module, and a generation module (if the answers are free-form text spans). 
\model{} achieves state-of-the-art performance on two benchmark conditional QA datasets, outperforming the previous state-of-the-art by 3-10 points.\footnote{Codes and data will be released upon the acceptance of this paper.}

\end{abstract}

\section{Introduction}


Recent work on QA has explored questions which have multiple possible answers, depending on conditions not explicitly given in the question \cite{min2020ambigqa,zhang2021situatedqa, dhingra2021timeaware,chen2021dataset}. For example, "when was the first Covid vaccine approved" has different answers for different countries, so answers must be completed with implicitly assumed conditions (e.g. Dec 20th, 2020 [in the US]").  In this work we focus on a more challenging task, in which answers rely on multiple conditions that logically interact.


An example is shown in Figure 1.  The span ``up to \$1200'' is an eligible answer, associated with two condition spans, ``you are partner \ldots of the deceased'' and ``you didn't claim other benefits''.  These two condition spans interact, as the answer ``up to \$1200'' is only valid if both conditions are satisfied.  We say that those conditions are a condition group and the logical type of the group is ``all'' (as witnessed by the span ``if both''). In addition to predicting eligible answers to the questions,  QA in this context additionally requires models to perform the two following tasks. First, it must understand the document well enough to parse it into eligible answers, condition groups, and logical types; second, it must identify which conditions are entailed by the question and scenario, which are contradicted, and which are not mentioned but are required to support an eligible answer. Given this a model can produce an answer together with a description of when that answer is supported --- i.e., by producing an eligible answer and associated conditions.


\begin{figure}
    \centering
    \includegraphics[width=0.98\textwidth]{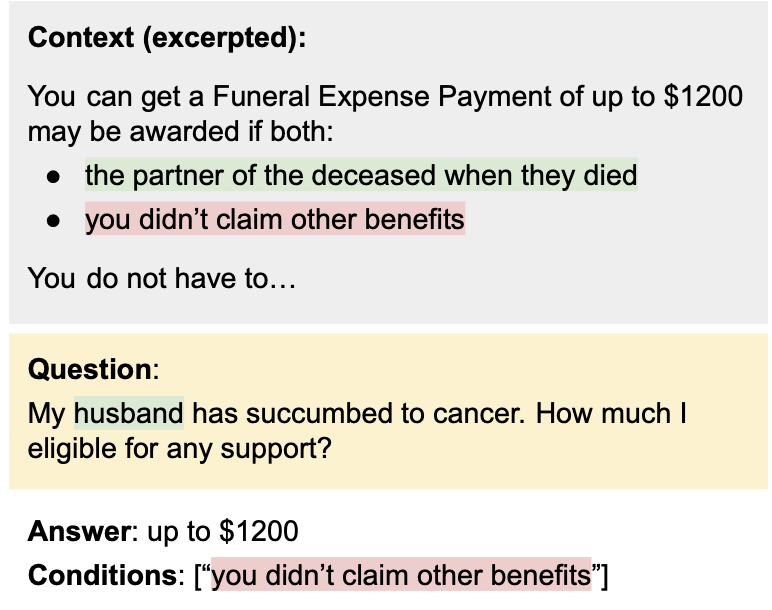}
    \vspace{-2pt}
    \caption{An example of reasoning over \textit{conditions}. The answer ``up to \$1200'' is only correct if ``both'' conditions are true. The scenario suggests that the user is a partner of the deceased but there's no evidence suggesting the condition ``you didn't claim any benefits''. Answering this question requires not only finding probable answers but also identifying unsatisfied conditions.}
    \label{fig:example}
\end{figure}

One of the challenges in this task is to perform logical reasoning within condition groups to determine the entailment status of conditions. The entailment status of a condition is affected by two factors: the entailment status of itself, i.e. whether it is satisfied or contradicted by the provided evidence, and the entailment status of other conditions in the same condition group. Similar tasks have been studied in \citet{ruletaker} where they constructed examples that contain groups of conditions which are either satisfied or contradicted by the evidences provided in the context (which is often called \textit{deductive reasoning} because all information needed to make a definite prediction is provided).
We do not make such assumption, but instead only provide evidences for a subset of conditions in the group and ask models to identify unsatisfied conditions that need to be checked for the answers. 
For example (Figure \ref{fig:example}), we say ``you didn't claim other benefits'' is an unsatisfied condition because it is required by the candidate answer ``up to \$1200'' but is not satisfied by the user's scenario. This task is called \textit{abductive reasoning}. Predicting unsatisfied conditions tests a model's ability in performing logical reasoning tasks, including understanding logical operations, determining the entailment status of conditions in the logical groups, and finally determining whether an eligible answer is correct. 

We propose the \model{} model to solve the QA task with multiple logically interacted conditions. \model{} contains two modules: an entailment module and a reasoning module. The entailment module takes a condition in the context with the question to predict its entailment status. The reasoning module takes the entailment module's outputs for all conditions to perform logical reasoning to identify unsatisfied conditions. If the answer is a free-form text span, \model{} additionally uses a generation module to generate the answer span. The entailment module, reasoning module, and generation module are jointly trained. \model{} shows excellent reasoning ability on a synthetic dataset and outperforms the previous state-of-the-art models on two Question Answering (QA) datasets, ConditionalQA and ShARC \cite{conditionalqa, sharc}, improving the state-of-the-art by 3-10 points on answer and unsatisfied condition prediction tasks.

\section{Related Work}

Models \cite{tensorlog,nql,emql,query2box,betaemb} have been developed for the deductive reasoning task with symbolic rules. Embedding-based methods \cite{emql,query2box,betaemb} first convert symbolic facts and rules to embeddings and then apply neural network layers on top to softly predict answers. Recent work in deductive reasoning focused on tasks where rules and facts are expressed in natural language \cite{talmor2020leap,rulebert,ruletaker,kassner2020pretrained}. Such tasks are more challenging because the model has to first understand the logic described in the natural language sentences before performing logical reasoning. 

Different from deductive reasoning, the QA task proposed in this paper provides a list of conditions that if true would support an answer. (This is also referred to as abductive reasoning.) The ConditionalQA and ShARC dataset \cite{conditionalqa,sharc} were proposed, where a question contains a user scenario that includes some background information that suggests the answer but is not enough to ensure its correctness. Similar examples were also seen in factual questions, e.g. AmbigQA \cite{min2020ambigqa}, where multiple answers are plausible given the facts asked in the question, but each answer is only correct under certain conditions. Answering such questions requires both finding the probable answers and identifying their underlying conditions. 


Very limited work has explored abductive reasoning for QA. Previous work \cite{emt,discern,dgm} on the ShARC \cite{sharc} dataset proposed to solve this problem by predicting a special label ``inquire'' if there was not enough information to make a definite prediction. The reasoning process was performed in the embedding space. Specifically, EMT and DISCERN \cite{emt,discern} computed an entailment vector for each condition and performed a weighted sum of those vectors to predict the final answer. DGM \cite{dgm} additionally introduced a GCN-based model to better represent the entailment vectors. Even though these models were able to predict the answer labels as ``inquire'' when there were unsatisfied conditions, none of them could predict which conditions needed to be further satisfied. 
Furthermore, they simply concatenated the full context and the question into a single input and encode it with a Transformer model with $O(N^2)$ complexity, making it not scalable to longer contexts.



\section{Model}
\begin{figure}
    \centering
    \includegraphics[width=1.0\textwidth]{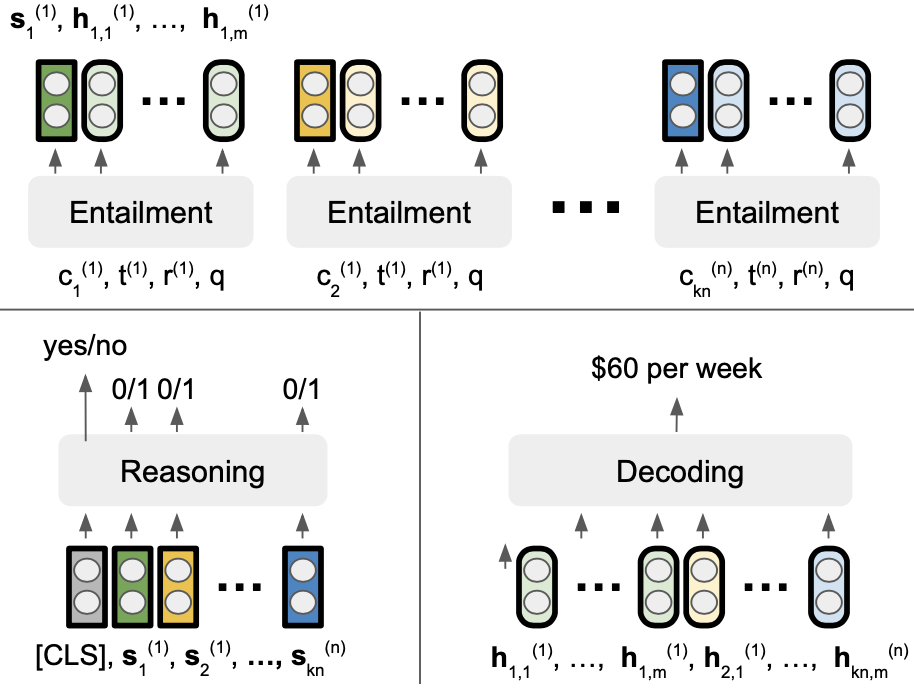}
    \caption{\textbf{\model{} Overview}. The top part shows the entailment module that independently encodes each condition $c_j^{(i)}$, its associated results $r^{(i)}$ and logical types $t^{(i)}$, and the question $q$. The entailment module outputs a condition embedding $\textbf{s}_j^{(i)}$ that will be input into the reasoning module (lower left) to predict the answer labels and determine unsatisfied conditions, and a token embedding $\textbf{h}_{k_i,p}^{(i)}$ that will be used by the decoding module (lower right) to generate answer spans (if the question has a free-form answer). All modules are jointly trained.
    \label{fig:treasoner_main}}
\end{figure}

\subsection{Task: QA with Conditions} \label{sec:model_input}
We study the task of QA with logically interacted conditions. The model learns to find candidate answers to the question from the context and additionally perform logical reasoning over the conditions to check whether the answers are eligible. If the answers require additional conditions to be satisfied, the model identifies these unsatisfied conditions as well.

The context used to answer the question contains \textit{results}, \textit{conditions}, and \textit{types}.
A \textit{result} is a sentence that contains the answer, e.g. ``You can get a Funeral Expense Payment of up to \$1200 ...''. Questions with yes/no answers also need result statements. For example, ``You don't need to pay taxes if you meet all of the following requirements: ...'' is the result statement for the question ``Do I need to pay taxes?''. A \textit{condition} describes a requirement that needs to be satisfied for a result to be applicable, e.g. ``physically or mentally disabled'' in Figure \ref{fig:example}. There could be multiple conditions for one result that interact under a logical \textit{type}. For example, ``if you're both:'' requires both conditions to be satisfied. In this project, we consider four logical types that are commonly seen in QA tasks:
\begin{itemize}
    \item ``all'': all conditions under this logical type should be satisfied in order to make the answer true. The logical type ``if you're both:'' in the example above is an example of this type.
    \item ``any'': only requires one of the conditions under the logical type ``any'' to be satisfied. For example, ``if you satisfy at least one of the following conditions.''. It doesn't matter whether other conditions have been satisfied, contradicted, or not mentioned in the question.
    \item ``required'': This is a special case of ``all'' / ``any'' when there is only one condition. Conditions with the logical type ``required'' must be satisfied. For example, ``you must ... to get an up to \$1200 payment.''
    \item ``optional'': Conditions have the type ``optional'' if they are not relevant to the question. For example, ``You will need to pay a \$30 processing fee if you apply online''.
    
\end{itemize}

Logical types are often not provided in real QA datasets, so they need to be inferred from the context. We discuss strategies to softly predict logical types to perform reasoning tasks for ConditionalQA in \S \ref{sec:exp_conditionalqa} and ShARC in \S \ref{sec:exp_sharc}. We will also discuss strategies to discover conditions and results from the context since they are often not labeled.

Formally, let the context $X$ consist of multiple results $r^{(1)}, \dots, r^{(n)}$, and the result $r^{(i)}$ be constrained by a group of conditions $\{c_1^{(i)}, \dots, c_{k_i}^{(i)}\}$ under the logical type $t^{(i)}$. We represent the context $X$ as a list of tuples $\{(\{c_1^{(1)}, \dots, c_{k_1}^{(1)}\}, r^{(1)}, t^{(1)})$, $(\{c_1^{(2)}, \dots, c_{k_2}^{(2)}\}, r^{(2)}, t^{(2)}), \dots\}$. We learn a model that operates on context $X$ and question $q$ to predict the answer $a$ with a list of unsatisfied conditions $\hat{C} = \{\hat{c}_1, \dots, \hat{c}_m\}$.\footnote{Some questions in the ConditionalQA dataset have multiple answers, but we do not handle these cases in this paper.}

\subsection{Model}\label{sec:model_main}

\model{} consists of an \textit{entailment module} and a \textit{reasoning module}. The entailment module checks whether a condition has been satisfied. Practically, it takes a condition, a result, and a question as its input, and outputs an learned embedding. Each condition will be encoded independently. Embeddings of all conditions will be passed to the reasoning module, which performs logical reasoning to predict the answer (if the answer is a multi-class label) as well as the unsatisfied conditions. 
In cases where answers are text spans, we apply an additional \textit{decoding module} to generate answer spans. All modules are jointly trained.

\noindent \textbf{Input} We independently encode each condition along with its associated result and the question. For the condition $c_j^{(i)}$ in $\{c_1^{(i)}, \dots, c_{k_i}^{(i)}\}$ with result $r^{(i)}$ under logical type $t^{(i)}$, we concatenate them and separate them with a special prefix.



\begin{equation} \label{eq:input}
\begin{split}
s_j^{(i)} = &~\textnormal{``condition:''} + ~c_j^{(i)} + \textnormal{``type:''} + ~t^{(i)} \\
          &+ ~\textnormal{``result:''} + ~r^{(i)} + \textnormal{``question:''} + ~q
\end{split}
\end{equation}

\noindent \textbf{Entailment Module} The entailment module encodes the concatenated input $s_j^{(i)}$ into vectors $\textbf{s}_j^{(i)}$. We initialize the parameters of the entailment module from pretrained LMs, which will be finetuned with other modules for the reasoning task. \begin{equation} \label{eq:entail}
    \textbf{s}_j^{(i)}, \textbf{h}_{j, 1}^{(i)}, \dots, \textbf{h}_{j, m}^{(i)} = \textnormal{Entail}(s_j^{(i)})
\end{equation}

The embedding of the first token of $s_j^{(i)}$ will be used as the embedding $\textbf{s}_j^{(i)}$ for condition $c_j^{(i)}$ and will be used by the reasoning module to predict the answer label (if the task is multi-class classification). $\textbf{h}_{j, 1}^{(i)}, \dots, \textbf{h}_{j, m}^{(i)}$ are the contextualized embeddings for the tokens in condition $s_j^{(i)}$. The token embeddings will not be used for reasoning but will be used for decoding if the answers are free-form answers.

At the entailment stage, each condition $s_j^{(i)}$ is encoded independently. The embedding output $\textbf{s}_j^{(i)}$ is expected to have information about the entailment state of the condition, the logical operation, and whether the result is relevant to the question. Encoding each condition independently also reduces the encoding complexity of all conditions in the passage from $O(C^2)$ to $O(C)$ where $C$ is the number of conditions in the provided passage, and thus enables handling longer context with hundreds of conditions. This encoding strategy is motivated by FiD \cite{fid}

\noindent\textbf{Reasoning Module} The reasoning module takes the embeddings from the entailment module and reasons over the conditions to predict the answer label (if the answer is a multi-class label). We use a Transformer model as our reasoner because the self attention mechanism allows conditions $\{s_1^{(i)}, \dots, s_{k_i}^{(i)}\}$ to attend to each other to perform the reasoning steps. It is crucial for reasoning because, for example, if one of the conditions is satisfied and the operation type is ``any'', then other conditions will be implicitly satisfied, regardless of their real entailment states.

We prepend a learned vector $\textbf{s}_0$ to the list of condition embeddings, which will be used as the \texttt{[CLS]} embedding to summarize the reasoning result. 
The output of the reasoning module, $\hat{\textbf{s}}_0, \hat{\textbf{s}}_1^{(1)}, \dots, \hat{\textbf{s}}_{k_n}^{(n)}$, will be used to predict the final label and unsatisfied conditions. 
Specifically, we use the first embedding $\hat{\textbf{s}}_0$ to predict the final answer label and use the subsequent embeddings $\hat{\textbf{s}}_1^{(1)}, \dots, \hat{\textbf{s}}_{k_n}^{(n)}$ to predict the entailment state of conditions as well as to identify unsatisfied conditions.
\begin{equation*}
    \begin{gathered}
        \hat{\textbf{s}}_0, \hat{\textbf{s}}_1^{(1)}, \dots, \hat{\textbf{s}}_{k_n}^{(n)} = \textnormal{Reason}(\textbf{s}_0, \textbf{s}_1^{(1)}, \dots, \textbf{s}_{k_n}^{(n)}) \\
        l_{\textnormal{label}} = \textnormal{softmax\_cross\_entropy}(\textbf{W}_l^T \hat{\textbf{s}}_0, \mathbb{I}_l) \\
        l_{\textnormal{cond}} = \textnormal{softmax\_cross\_entropy}(\textbf{W}_c^T \hat{\textbf{s}}_{j}^{(i)}, \mathbb{I}_{c})
    \end{gathered}
\end{equation*}


\noindent where $\mathbb{I}_{l}$ and $\mathbb{I}_{c}$ are one-hot vectors for the class labels. The number of label classes are task-dependent, but in most cases, the final labels $\mathbb{I}_{l}$ are ``yes'', ``no'', and ``irrelevant''. The condition labels are ``entailed'', ``contradicted'', ``not mentioned'', ``implied'', and ``to check''. The first three classes are as they are named. The class ``implied'' means the entailment state of this condition is implied by other conditions with the same result, e.g. if one of the conditions with the logical type ``any'' has been satisfied, the rest of conditions are automatically ``implied''. The class ``to check'' means it is an unsatisfied condition. It is important to note that the condition loss $l_{\textnormal{cond}}$ is an auxiliary loss and may not exist (or only exist for a subset of conditions) in real datasets. 

For questions that have free-form answers, e.g. ``up to \$1200'', the answers will be generated from the decoding module discussed in the next section. We will not supervise their class labels in training and can safely ignore the predicted label in testing. In this case, only the predictions of the unsatisfied conditions will be kept. On the contrary, for questions that have multi-class answers, the reasoning module is trained to predict the correct label while the decoding module (discussed next) is trained to generate a special token \texttt{[MULTI]}.




\noindent\textbf{Decoding Module} The decoding module takes token embeddings for all conditions $\textbf{h}_{1, 1}^{(1)}, \dots, \textbf{h}_{k_n, m}^{(n)}$ to generate the answer span. This module is mostly used when the final answer is a text span. If the answer is a multi-class label, the decoding module should simply generate a special token \texttt{[BIN]}.

We adopt the decoding strategy proposed by FiD \cite{fid} with the T5 architecture \cite{t5}\footnote{The T5 encoder is used for the entailment module}, i.e. the token embeddings are concatenated for decoding even though the they are generated independently for each condition. The generation task is trained with teacher forcing. We do not write out the explicit expression for the teacher forcing decoding loss $l_{\textnormal{decode}}$ here. Please refer to the T5 paper \cite{t5} for more information. The decoded tokens $\hat{a}$ are taken as the predicted answer span.
\begin{equation} \label{eq:decoder}
    \hat{a} = \textnormal{Decode}(\textbf{h}_{1, 1}^{(1)}, \dots, \textbf{h}_{k_n, m}^{(n)})
\end{equation}

\noindent \textbf{Loss Function} We jointly train the entailment module and reasoning module. We provide intermediate supervision on the entailment state of each condition, i.e. $\textbf{s}_{j}^{(i)}$, if they are available. The final loss function is the sum of the answer loss $l_{\textnormal{label}}$ and the condition entailment loss $l_{\textnormal{cond}}$.
$$
l = l_{\textnormal{label}} + l_{\textnormal{cond}}
$$

If the answers contain text spans, we jointly train the decoding module as well. The loss function is the sum of the three losses:
$$
l = l_{\textnormal{label}} + l_{\textnormal{cond}} + l_{\textnormal{decode}}
$$

\noindent\textbf{Pretrained Checkpoints} The entailment module and decoding module (if any) load pretrained LM checkpoints and finetune the parameters for downstream tasks. For the dataset that has both multi-class answers and free-form answers, we initialize the entailment module and decoding module with the pretrained T5 encoder and decoder \cite{t5}. For a dataset that only has multi-class answers, the decoding module is not needed, so only the entailment module will be initialized. The entailment module can be initialized with T5 (encoder only) or any other pretrained LMs, e.g. BERT, RoBERTa, ELECTRA, BART, \cite{bert,roberta,electra,bart}, etc. We use ELECTRA for our entailment module if the decoding module is not needed.


The reasoning module is randomly initialized and jointly trained with the entailment and decoding modules. The number of Transformer layers for the reasoning module is a hyper-parameter. We choose the number of layers $l=3$ or $l=4$. Please see \S \ref{sec:condent_abl} for ablation study on the number of Transformer layers for the reasoning task.

\section{Experiments}
We experiment \model{} with a synthetic dataset, \mydataset{}, and two QA datasets, ConditionalQA \cite{conditionalqa} and ShARC \cite{sharc}, that require reasoning over conditions to predict the answers.

\subsection{\mydataset{}} 
\begin{table}[t]
\small
\centering
\begin{tabular}{ll}
\toprule
(Template) & \\
\multicolumn{2}{l}{\begin{tabular}[c]{@{}l@{}}Context: If all (A, B), then U. \\ ~~~~~~~~~~~~~~~If any (not C, D), then V.\\ Facts: a, c, not d.\\ Question: Is u correct?\\ Label: entailed, if B\end{tabular}} \\
\midrule
(Variables) & \\
A: Aged 59 1/2 or older.                                                                     & a: Tom is 65 years old.                                                                     \\
                           B: Employed for two years.                                                          & b: NOT\_USED                                                                                \\
                           C: Has two children                                                                          & c: He has two sons.                                                          \\
                           D: Has not applied before.                                                                   & not d: Rejected last year.                                                          \\
                           U: Get at least \$60 a week                                                                  & u: Eligible for \$60 a week.                                                                \\
                           V: Waive the application fees                                                                & v: NOT\_USED                                                                               \\
\bottomrule
\end{tabular}
\caption{\label{tab:condnli_example}An example of \mydataset{}. Variables $A$, $B$, $\dots$ and $U$, $V$, $\dots$ represent the conditions and premises. Variables $a$, $b$, $\dots$ represent the known facts. $u$ is the question. Each pair of variables, e.g. ($A$, $a$), is instantiated with an NLI example. }
\end{table}

The \mydataset{} dataset is constructed from the existing Natural Language Inference (NLI) dataset, MultiNLI \cite{multinli}. In the original NLI dataset, an example has a premise, a hypothesis, and an entailment label, e.g. ``entailed'', ``contradicted'' or ``neutral''. 
To construct the \mydataset{} dataset, we treat the premise as context and the hypothesis as question, and make a few additional changes.
First, each premise is paired with a list of conditions $c_j$'s that interact under a logical type $t$. 
Second, a context contains multiple premises, but only one of the premises are relevant to the hypothesis.\footnote{In some examples, none of the premises is relevant to the hypothesis. Such examples will be labeled as ``Irrelevant''.} The model should first identify the relevant premise and then check their associated conditions to predict the final labels. 
Third, each example is provided an additional list of known facts for checking the entailment state of the conditions. All premise, hypothesis, conditions, and facts are obtained from the MultiNLI dataset \cite{multinli}. Table \ref{tab:condnli_example} gives an example in \mydataset{}.



\subsubsection{Dataset Construction} \label{sec:mydataset}
We first construct templates for the \mydataset{} examples and then replace the variables in the template with real NLI examples. 


\noindent \textbf{Construct Templates} We use capital letter variables $U$, $V$, $\dots$ to represent the premises in the context, and $A$, $B$, $\dots$ to represent the conditions. As discussed above, a premise is paired with a list of conditions and a logical operation. We express the relationship between the premise and the conditions with the statement ``if ... then ...''. For example in Table \ref{tab:condnli_example}, we say ``If all ($A$, $B$), then $U$'' to represent that the premise $U$ has the conditions $A$ and $B$, and the logical operation ``all''.

Since the question is only about one of the premises in the context, we randomly sample a premise, e.g. $U$, and take its corresponding hypothesis $u$ as the question. With the question $u$, only the conditions of the premise $U$ need to be satisfied.

We also provide a list of facts that are used to check the entailment state of the conditions. To construct the facts, we randomly sample a subset of the conditions from the context, e.g. \{$A$, $C$, $D$\}, and take the facts of the selected conditions, e.g. \{$a$, $c$, $d$\}. Furthermore, we randomly add the term ``not'' to a fact, e.g. not $d$, to indicate that the fact $d$ contradicts with its condition $D$.

With the question, e.g. $u$, and the list of facts e.g. \{$a$, $c$, not $d$\}, we can infer the answer label and identify unsatisfied conditions. We keep the label ``entailed'', ``contradicted'', and ``neutral'', and add an additional label ``irrelevant'' if none of the premise in the context is relevant to the document.

\noindent \textbf{Generate Examples} For a templates with variables $A$, $B$, $U$, $V$, $\dots$, $a$, $b$, $u$, $v$, $\dots$, we instantiate the variables with NLI examples to get the real data. We use the premises of original NLI examples for premises or conditions, i.e. capital letter variables, and the hypothesis for question and facts, i.e. lower-case variables. Note that sampling requires matching the entailment state of conditions, e.g. ``not $d$'' requires sampling from NLI examples with the original label ``contradict''. 

We restrict the number of conditions in the context to 6 and randomly generate 65 distinct templates.\footnote{Restricting the number of conditions is only for the purpose of reducing training complexity. The experiment in Figure \ref{fig:condent_abl} (left) shows the model's capability of generalizing to more conditions.} During training, we randomly pick a template and instantiate it with NLI examples to generate real training examples. This random generation process enables creating (almost) unlimited amount of training data. We randomly generate another 5000 examples for development and testing.


\subsubsection{Results}\label{sec:condent_abl}
Previous work \cite{ruletaker} showed that Transformer-based Language Models, e.g. RoBERTa \cite{roberta}, have the ability to reason over multiple conditions to answer the reasoning question in the deductive reasoning setting, e.g. ``if A and B then C'' with facts on conditions A and B provided. We replace RoBERTa with FiD with T5 here to handle long contexts. FiD is trained to generate the answer labels and the list of unsatisfied conditions. To simplify the generation task, we prepend a condition id to each condition and let the model generate the condition id instead.

We train the model on two types of input, one using templates with variables in letters, and the other using examples where variables are instantiated with real NLI examples.

\begin{table}[]
\small
\centering
\begin{tabular}{lcc}
\toprule
             & Label & Conditions \\
             & (acc) & (F1) \\
\midrule
(template) & & \\
~~FiD (concat) & 99.8    & 98.7           \\
~~FiD (\model{})   & 99.6    & 99.2         \\
~~\model{}         & 99.8  & 99.2          \\
\midrule
(with NLI) & & \\
~~FiD (concat) & 85.6    & 80.4           \\
~~FiD (\model{})   & 86.7    & 82.8         \\
~~\model{}         & 95.0  & 91.3          \\
\bottomrule
\end{tabular}
\caption{\label{tab:condent_result} Experiment results on the \mydataset{} dataset in label accuracy and condition F1. FiD (concat) is run on the input that concatenates the question and context and is then chunked into smaller pieces. FiD (\model{}) use the same input as \model{}. ``(template)'' directly uses the templates with variable letters as inputs, while ``(with NLI)'' uses the examples that are instantiated with real NLI examples.}
\end{table}

\noindent \textbf{Main Results} The experiment results are shown in Table \ref{tab:condent_result}. We measure both the accuracy of label prediction and the F1 of unsatisfied conditions. The results show that a plain Transformer-based sequence-to-sequence model (FiD) performs the logical reasoning task reasonably well if the context is simple, i.e. using the template with variables $A$, $B$, $\dots$ as inputs. However, the FiD performs significantly worse on examples with real NLI examples. \model{} still performs well on the \mydataset{} dataset with NLI examples.

\begin{figure}[h]

\begin{subfigure}{0.49\textwidth}
\includegraphics[width=0.98\linewidth]{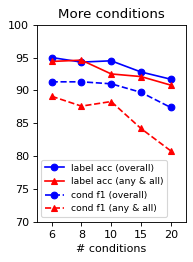} 
\label{fig:subim1}
\end{subfigure}
\begin{subfigure}{0.49\textwidth}
\includegraphics[width=0.98\linewidth]{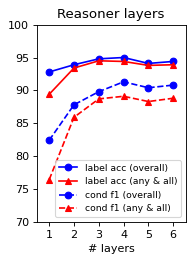}
\label{fig:subim2}
\end{subfigure}

\caption{\label{fig:condent_abl} Left: Generalization results of reasoning over more conditions. Right: Results on the ablated model with different numbers of Transformer layers in the reasoning module. We report both label accuracy and F1 of unsatisfied conditions. ``any \& all'' indicates that examples only have two types of logical operation: ``any'' and ``all''.}

\end{figure}

\noindent \textbf{Generalization to More Conditions} The \model{} is trained on templates with 6 conditions or fewer. To test \model{}'s ability to generalize to more conditions, we take a trained model and test it on the examples with more than 6 conditions. Figure \ref{fig:condent_abl} (left) shows the change of performance in both label classification and unsatisfied condition prediction tasks as the number of conditions increase.\footnote{``any \& all'' indicates that the context only contains conditions under the logical operation ``any'' or ``all''.} We observe more decrease in performance in predicting unsatisfied conditions (probably because more conditions are unsatisfied), but it is still reasonable with 20 conditions.

\noindent \textbf{Number of Reasoning Layers} We additionally experiment with different numbers of layers in the reasoner module. The results are shown in Figure \ref{fig:condent_abl} (right). The Transformer-based reasoner module needs at least 3 layers to perform the reasoning task, especially for predicting unsatisfied conditions.

\subsection{ConditionalQA} \label{sec:exp_conditionalqa}
\begin{table*}[]
\small
\centering
\begin{tabular}{lcccccccc}
\toprule
                          & \multicolumn{2}{c}{Yes / No}                & \multicolumn{2}{c}{Extractive}              & \multicolumn{2}{c}{Conditional}             & \multicolumn{2}{c}{Overall}                 \\
                          & EM / F1               & w/ conds             & EM / F1               & w/ conds             & EM / F1               & w/ conds            & EM / F1               & w/ conds             \\ \midrule
majority                  & 62.2 / 62.2          & 42.8 / 42.8          & -- / --              & -- / --              & -- / --              & -- / --              & -- / --              & -- / --              \\
ETC                       & 63.1 / 63.1          & 47.5 / 47.5          & 8.9 / 17.3           & 6.9 / 14.6           & 39.4 / 41.8          & 2.5 / 3.4            & 35.6 / 39.8          & 26.9 / 30.8          \\
DocHopper                 & 64.9 / 64.9          & 49.1 / 49.1          & 17.8 / 26.7          & 15.5 / 23.6          & 42.0 / 46.4          & 3.1 / 3.8            & 40.6 / 45.2          & 31.9 / 36.0          \\
FiD                       & 64.2 / 64.2          & 48.0 / 48.0          & 25.2 / 37.8          & 22.5 / 33.4          & 45.2 / 49.7          & 4.7 / 5.8            & 44.4 / 50.8          & 35.0 / 40.6          \\
\model{} & \textbf{73.2 / 73.2} & \textbf{54.7 / 54.7} & \textbf{34.4 / 48.6} & \textbf{30.3 / 43.1} & \textbf{51.6 / 56.0} & \textbf{12.5 / 14.4} & \textbf{57.2 / 63.5} & \textbf{46.1 / 51.9} \\ 
\bottomrule
\end{tabular}
\caption{\label{tab:conditionalqa_result} Experimental results on ConditionalQA (EM / F1). The ``EM/F1'' columns reports the original EM/F1 metrics that are only evaluated on the answer span. The ``w/ conds'' is the conditional EM/F1 metric discussed in \S \ref{sec:conditionalqa_eval}. Numbers of the baseline models are obtained from \citet{conditionalqa}.}
\end{table*}

In the second experiment, we run \model{} on a real question answering (QA) dataset, ConditionalQA \cite{conditionalqa}, that requires reasoning over long documents with much more conditions and more complex logical operations stated in natural language. 

\subsubsection{Task}
ConditionalQA is challenging because it requires the model to accurately locate relevant results and conditions from longer documents. Previous models, e.g. RuleTaker, DGM \cite{ruletaker, dgm} concatenate the inputs into a long sequence and then compute cross-attention over the concatenate input. The length of the input is constrained by the $O(N^2)$ complexity. Even if we adopt the Fusion-in-Decoder \cite{fid} strategy to handle long sequences, performance is still limited (see Table \ref{tab:condent_result}).

Another challenge of the ConditionalQA dataset is to identify the logical operation for the conditions. As it is shown in the example in Figure \ref{fig:example}, the model should predict the logical operation ``all'' from the statement that ``if you're both:''. One could possibly provide intermediate supervision to learn to predict the logical operations. However, such labels are not provided in the dataset and it is hard to find distant supervision labels. \model{} encodes the logical operation in the condition's embeddings $\textbf{s}_j^{(i)}$ (Eq. \ref{eq:entail}) and does not need additional supervision for the logical operation.

Furthermore, different from the \mydataset{} and ShARC datasets (see \S \ref{sec:exp_sharc}), the ConditionalQA dataset contains questions with both yes/no answers and free-form answer spans. We apply the decoder module on the token embeddings $\textbf{h}_{1, 1}^{(1)}, \dots, \textbf{h}_{k_n, m}^{(n)}$ to generate the final answer spans (Eq. \ref{eq:decoder}).

\subsubsection{Data Preparation}


Examples in the ConditionalQA dataset provide a parsed web page as context, a question, and a user scenario that describes some relevant information about the question. 
We parse the provided context into the format that contains a list of tuples $\{(\{c_1^{(1)}, \dots, c_{k_1}^{(1)}\}, r^{(1)}, t^{(1)}), \dots\}$ as in \S \ref{sec:model_input}. 

The context in ConditionalQA is provided as a list of HTML elements. We treat each element at the leaf of the DOM tree as a condition, and all its parents (from its direct parent to the root) as the result. Conditions under the same parent are considered to be in the same list $\{c_1^{(i)}, \dots, c_{k_i}^{(i)}\}$. As discussed before, the logical operations $t^{(i)}$ need to be inferred from the context. We drop the field ``type:'' in the input in Eq. \ref{eq:input} and ask the model to discover it from the context and implicitly encode it into the condition embeddings $\textbf{s}_j^{(i)}$. The question $q$ is the combination of the question and scenario.

\subsubsection{Evaluation}\label{sec:conditionalqa_eval}

The predictions are evaluated using two sets of metrics: EM/F1 and conditional EM/F1. EM/F1 are the traditional metrics that measures the predicted answer spans. The ConditionalQA dataset introduced another metric, \textit{conditional EM/F1}, that jointly measures the accuracy of the answer span and the unsatisfied conditions. Please refer to the ConditionalQA paper \cite{conditionalqa} for more information. Briefly, the conditional EM/F1 is the product of the original answer EM/F1 and the F1 of the predicted unsatisfied conditions. The conditional EM/F1 is 1.0 if and only if the predicted answer span is correct and all unsatisfied conditions are found. If there's no unsatisfied condition, the model should predict an empty set.

\subsubsection{Results}

We compare \model{} with a few baseline models, including ETC (in a pipeline) \cite{etc}, DocHopper \cite{dochopper}, and Fusion-in-Decoder (FiD) \cite{fid}. The ETC pipeline first extracts possible answers from the context and then takes the question and extracted answers as input to find unsatisfied conditions. The answer extraction model and the condition prediction model are trained separately. DocHopper is a multi-hop attention system that iteratively attends to sentences to jointly predict the answers and unsatisfied conditions. The iterative process in DocHopper is updated in the embedding space so it is end-to-end differentiable. FiD is a encoder-decoder model based on T5. FiD improves over T5 by proposing to split long input sequences into short sequences, encode the short sequences independently, and jointly decode over all encoded embeddings to generate the outputs. For the ConditionalQA dataset, we train the FiD model to generate the answers followed by the list of unsatisfied conditions.

\noindent\textbf{Main Results} The experimental results are presented in Table \ref{tab:conditionalqa_result}. \model{} achieves the state-of-the-art on both yes/no and extractive questions. \model{} also significantly outperforms all the baselines on the questions with conditional answers with 166\% and 148\% relative improvement in the conditional EM/F1 metrics (w/ conds). 


\noindent\textbf{Condition Accuracy} Since there's not a metric that directly measure the quality of predicted conditions, we additionally report the F1 of the predicted unsatisfied conditions (Table \ref{tab:condent_result}). The best baseline models, FiD, rarely predicts any conditions. This is likely because only a subset of the questions have unsatisfied conditions. Even though we train the FiD model only on the subset of questions that have conditional answers, its performance slightly improves but is still much lower than \model{} by 16.5 points in condition F1.

\begin{table}[]
\small
\centering
\begin{tabular}{lcc}
\toprule
                       & \begin{tabular}[c]{@{}c@{}}Answer\\ (w/ conds)\end{tabular} & \begin{tabular}[c]{@{}c@{}}Conditions\\ (P / R / F1)\end{tabular} \\
\midrule
FiD                    & 3.2 / 4.6                                                        & 98.3 / 2.6 / 2.7                                                  \\
FiD (conditional only) & 6.8 / 7.4                                                        & 12.8 / 63.0 / 21.3                                                                  \\
\model{}                  &  \textbf{10.6 / 12.2}                                               & \textbf{34.4 / 40.4 / 37.8}                                                                 \\
\bottomrule
\end{tabular}
\caption{\label{tab:conditionaqa_result_conditions} Experimental results on the subset of questions in ConditionalQA (dev) that has conditional answers. Accuracy for the answers is evaluated using the conditional EM/F1 (w/ conds) metrics defined by \citet{conditionalqa}. Conditions are evaluated in precision, recall and F1.}
\end{table}

\subsection{ShARC}\label{sec:exp_sharc}

We additionally experiment \model{} with the ShARC \cite{sharc} dataset. The ShARC dataset examples have shorter context, usually a few sentences or a short passage, but the logical operations between conditions are more complex, as is discussed below.


\subsubsection{Task}

The ShARC dataset has two subtasks: Decision Making and Question Generation. The decision making task asks the model to predict one of the following labels as the answer: ``yes'', ``no'', ``inquire'', and ``irrelevant''. The label ``inquire'' means that information provided by the question is not enough to make a definite prediction, i.e. there are unsatisfied conditions. In this case, the model should perform the Question Generation task to generate a followup question to clarify the unsatisfied conditions. The decision making task evaluates the predicted labels using micro and macro accuracy. The question generation task evaluates the BLEU scores of the generated question with the ground truth annotation. Note that some example could have multiple unsatisfied conditions, but only one of them will be annotated as ground truth followup question.\footnote{To mitigate this issue in evaluation, we run an additional evaluation that measures the F1 of the predicted unsatisfied conditions. Please see results in Table \ref{tab:sharc_cond}.}

\subsubsection{Data Preparation}
Different from ConditionalQA, where each sentence in the context is treated as a condition, conditions in the ShARC dataset are shorter and are sometimes short phrases (sub-sentence). For example, the context ``If you are a female Vietnam Veteran with a child who has a birth defect, you are eligible for ...'' contains two conditions, ``If you are a female Vietnam Veteran'' and ``with a child who has a birth defect''.\footnote{It is arguable that this could be generally treated as one condition, but it is treated as two conditions with the logical operator ``all'' in the ShARC dataset.} In order to handle sub-sentence conditions, we follow the strategy proposed in two of the baseline models, DISCERN \cite{discern} and DGM \cite{dgm}, that split a sentence into EDUs (Elementary Discourse Units) using a pretrained discourse segmentation model \cite{li2018segbot}. The discourse segmentation model returns a list of sub-sentences, each considered as a condition.

While we could treat each condition independently as we did previously for other datasets, the segmented EDUs are different in that they are not full sentences and may not retain their semantic meaning. Thus, we jointly encode all EDUs $s_j^{(i)}$ as a single passage and select embeddings at specific tokens in the sentence as the condition embeddings $\textbf{s}_j^{(i)}$. We construct the input $s$ for the entailment module as followed. 
\begin{equation*} 
\begin{split}
s = &~\textnormal{``condition:''} + ~c_1^{(1)} + \dots \\
+ &~\textnormal{``condition:''} + ~c_{k_n}^{(n)} \\
    + &~\textnormal{``question:''} + ~q
\end{split}
\end{equation*}

Similar to ConditionalQA, we drop the ``type:'' argument because the logical operation is not provided and needs to be inferred from the context. We additionally drop the argument ``result:'' and let the model to attend to one or more of the EDUs (with the argument ``condition:'') as the result. The input $s$ is used to compute the condition embeddings 
for the EDUs.
The condition embedding $\textbf{s}_j^{(i)}$ for the EDU $c_j^{(i)}$ is the embedding at its preceding token ``condition:''.

$$
\textbf{s}_1^{(1)}, \dots, \textbf{s}_{k_n}^{(n)} = \textnormal{Entail} (s)
$$

For the question generation task, we use the same input $s$ as in the decision making task, except that we replace the prefix ``condition:'' with  ``unsatisfied condition:'' for the conditions that are classified as unsatisfied conditions. We fine-tune a pretrained T5 model for the question generation task.


\subsubsection{Results}
\begin{table}[t]
\small
\centering
\begin{tabular}{lcc}
\toprule
        & Decision & Question \\
        & (micro / macro)            & (BLEU1 / BLEU4)              \\
\midrule
CM      & 61.9 / 68.9             & 54.4 / 34.4               \\
BERTQA  & 63.6             / 70.8             & 46.2               / 36.3               \\
UcraNet & 65.1             / 71.2             & 60.5               / 46.1               \\
Bison   & 66.9             / 71.6             & 58.8               / 44.3               \\
E3      & 67.7             / 73.3             & 54.1               / 38.7               \\
EMT     & 69.1             / 74.6             & 63.9               / 49.5               \\
DISCERN & 73.2             / 78.3             & 64.0               / 49.1               \\
DGM     & 77.4             / 81.2             & 63.3               / 48.4               \\
\model{}   & \textbf{80.4}    / \textbf{83.9}    & \textbf{71.5}      / \textbf{58.0}     \\
\bottomrule
\end{tabular}
\caption{\label{tab:sharc} Experimental results on the ShARC dataset. Numbers for the baseline models \cite{sharc, bertqa, ucranet, bison, emt, discern, dgm} are borrowed from \citet{dgm}. }
\end{table}

\noindent\textbf{Main Results} We compare \model{} to a few strong baseline models, including the previous state-of-the-art model, e.g. DISCERN and DGM \cite{discern,dgm}. Different from the baseline models, which use separate models for label classification and unsatisfied condition prediction, \model{} performs both tasks jointly.\footnote{Previous models, e.g. DISCERN and DGM, additionally use a generation model to paraphrase the unsatisfied conditions into questions, similar to our generation process with T5.} The results are shown in Table \ref{tab:sharc}. \model{} outperforms the previous baselines by 3 points on the classification task and more than 8 points on the question generation task. 

\noindent\textbf{Condition Accuracy} One problem with the current question generation task is that the ground-truth question only asks about one of the unsatisfied conditions, even though there could be multiple unsatisfied conditions. To further evaluate \model{}'s performance in predicting unsatisfied conditions, we manually annotate the logical operations in 20 passages that have more than one condition (857 data total),\footnote{Each passage in ShARC has 32.9 data on average.} and use the annotated logical operations to find all unsatisfied conditions. We report the F1 of the predicted unsatisfied conditions (see Table \ref{tab:sharc_cond}). Compared to the baselines \cite{discern, dgm}, \model{} improves the F1 by 11.4.

\begin{table}[t]
\small
\centering
\begin{tabular}{lccc}
\toprule
      & Decision             & Question             & Condition     \\
      & (micro / macro)      & (BLEU1 / 4)      & (F1)            \\
\midrule
DISCERN  & 74.9 / 79.8  & 65.7 / 52.4 & 55.3 \\
DGM   & 78.6 / 82.2          & \textbf{71.8 / 60.2} & 57.8          \\
\model{} & \textbf{79.8 / 83.5} & \textbf{71.7 / 60.4} & \textbf{69.2} \\
\bottomrule
\end{tabular}
\caption{\label{tab:sharc_cond} Experiment results on the ShARC dataset (dev) compared to the baselines, DISCERN and DGM \cite{discern, dgm}. The Condition (F1) number is obtained by reruning their open-sourced codes.}
\end{table}

\noindent\textbf{Label Accuracy v.s. Conditions} We additionally measure the accuracy versus the number of conditions in the context. We consider the number of all followup questions on each context as its number of conditions. Results in Table \ref{tab:sharc_acc_vs_cond} show that the improvement in \model{}'s performance over the previous state-of-the-art model (DGM) mostly come from questions that have more than one conditions. 
\begin{table}[t!]
\centering
\small
\begin{tabular}{lcccc}
\toprule
\# conditions & 1    & 2    & 3    & 4    \\
\midrule
DGM      & 90.4 & 70.3 & 80.0 & 73.4 \\
\model{}           & 90.3 & 72.7 & 80.6 & 75.2 \\
\midrule
\textit{diff}          & -0.1 & 2.4  & 0.6  & 1.8 \\
\bottomrule
\end{tabular}
\caption{\label{tab:sharc_acc_vs_cond}Ablation study on the label accuracy vs. the number of conditions in the context. Numbers of DGM \cite{dgm} is obtained by reruning their open-sourced codes.}
\end{table}

\section{Conclusion}

We study the problem of QA with answers that are constrained by a list of conditions that interact with each other under logical operations, such as ``any'' or ``all''. We propose a system, \model{}, that contains an entailment module to check the entailment status of conditions and a jointly trained reasoning module that performs the logical reasoning to predict the final answers and the unsatisfied conditions. \model{} shows excellent reasoning ability, and can easily generalize to more conditions on a synthetic dataset \mydataset{}. Furthermore, \model{} achieves the state-of-the-art performance on two challenging question answering datasets ConditionalQA and ShARC \cite{conditionalqa,sharc}.


\end{document}